\documentclass[10pt,twocolumn,letterpaper]{article}
\usepackage[pagenumbers]{iccv}    
\usepackage{xcolor}
\usepackage{arydshln}
\usepackage{booktabs}
\usepackage{pifont}
\usepackage{float}
\usepackage{algorithm}
\usepackage{algorithmicx}
\usepackage{multirow}
\usepackage{adjustbox}
\usepackage[noend]{algpseudocode}

\newcommand{\x}{{\mathbf{x}}}
\newcommand{\xhq}{{\x_\text{HQ}}}
\newcommand{\xhqhat}{{\hat{\x}_\text{HQ}}}
\newcommand{\xlq}{{\x_\text{LQ}}}
\newcommand{\D}{{\mathcal{D}}}
\newcommand{\R}{{\mathcal{R}}}
\newcommand{\supp}{{\textbf{\textcolor{blue}{Supplementary Material}}}}
\newcommand{\best}[1]{\textbf{\textcolor{red}{#1}}}
\newcommand{\sbest}[1]{\underline{\textcolor{blue}{#1}}}

\definecolor{iccvblue}{rgb}{0.21,0.49,0.74}
\usepackage[pagebackref,breaklinks,colorlinks,allcolors=iccvblue]{hyperref}
\def\paperID{} 
\def\confName{}
\def\confYear{2025}
\title{Multi-Agent Image Restoration}

\author{
Xu Jiang$^{*}$ \qquad
\href{https://github.com/cvsym}{Gehui Li}$^{*}$ \qquad
\href{https://scholar.google.com/citations?user=aZDNm98AAAAJ}{Bin Chen}$^{*}$ \qquad
\href{https://jianzhang.tech/}{Jian Zhang}$^{\dagger}$ \qquad \\
School of Electronic and Computer Engineering, Peking University \qquad\\
{\tt\footnotesize xujiang2642081986@gmail.com}\qquad
{\tt\footnotesize \{chenbin,ligehui921\}@stu.pku.edu.cn} \qquad
{\tt\footnotesize \ zhangjian.sz@pku.edu.cn}}

\begin{document}
\maketitle

\begin{abstract}
Image restoration (IR) is challenging due to the complexity of real-world degradations. While many specialized and all-in-one IR models have been developed, they fail to effectively handle complex, mixed degradations. Recent agentic methods RestoreAgent and AgenticIR leverage intelligent, autonomous workflows to alleviate this issue, yet they suffer from suboptimal results and inefficiency due to their resource-intensive finetunings, and ineffective searches and tool execution trials for satisfactory outputs. In this paper, we propose \textbf{MAIR}, a novel \textbf{M}ulti-\textbf{A}gent approach for complex \textbf{IR} problems. We introduce a real-world degradation prior, categorizing degradations into three types: (1) scene, (2) imaging, and (3) compression, which are observed to occur sequentially in real world, and reverse them in the opposite order. Built upon this three-stage restoration framework, MAIR emulates a team of collaborative human specialists, including a ``scheduler'' for overall planning and multiple ``experts'' dedicated to specific degradations. This design minimizes search space and trial efforts, improving image quality while reducing inference costs. In addition, a registry mechanism is introduced to enable easy integration of new tools. Experiments on both synthetic and real-world datasets show that proposed MAIR achieves competitive performance and improved efficiency over the previous agentic IR system. Code and models will be made available.
\end{abstract}

\renewcommand{\thefootnote}{}
\footnote{$^*$Equal Contribution. $^\dagger$Corresponding Author.}

\begin{figure}[!t]
\vspace{-20pt}
\centering
\includegraphics[width=\linewidth]{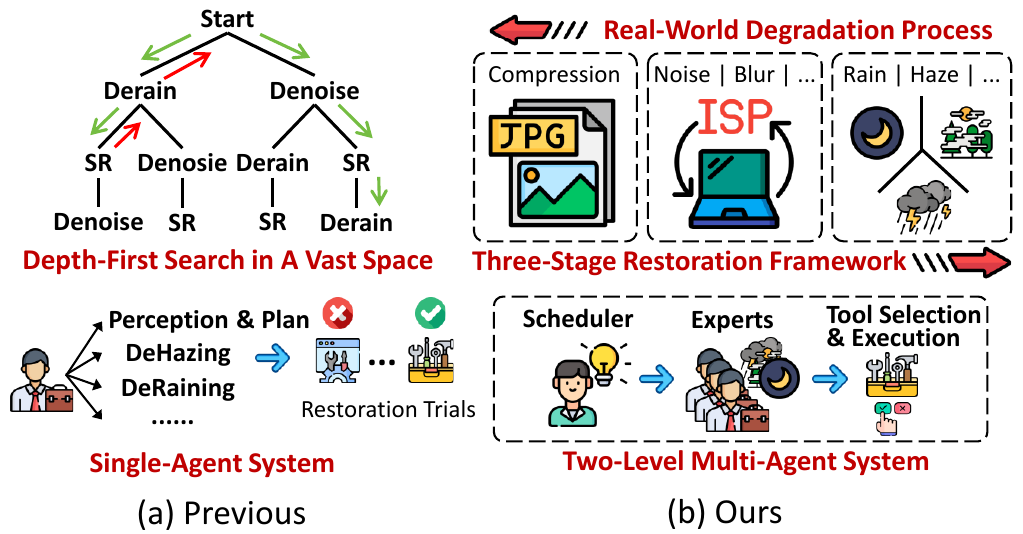}
\vspace{-17.5pt}
\caption{\textbf{Comparison between our proposed MAIR and typical agentic IR approaches.} \textbf{(a)} The state-of-the-art method AgenticIR \cite{agenticir} employs a single agent for perception, planning, restoration, \etc, suffering from resource-intensive searches and trials for degradation removal. \textbf{(b)} We decompose the complex IR problem into manageable sub-tasks and address them using multiple collaborative agents, under our proposed three-stage restoration framework, achieving improved performance, efficiency, and flexibility.}
\label{fig:teaser}
\vspace{-10pt}
\end{figure}

\vspace{-25pt}
\section{Introduction}
\label{sec:intro}
Image restoration (IR) is a long-standing, challenging problem in computer vision. It aims to reconstruct high-quality (HQ) original images from low-quality (LQ) degraded ones. Traditional deep IR networks \cite{wu2020unpaired,zhang2022idr,yu2024scaling,lin2024diffbir,wu2024seesr,wang2021real,chen2024adversarial,li2023fearless,yang2024difflle} are typically designed for specific IR tasks, focusing on single degradations such as rain, haze, noise, blur, and JPEG compression. However, in real-world scenarios, images often suffer from multiple degradations that can interact with each other, significantly increasing the complexity of IR.

To address this, researchers have developed All-in-One (AiO) IR methods \cite{park2023all,li2022all,jiang2024autodir,potlapalli2024promptir,conde2024instructir,kong2024towards,luo2023controlling,ai2024multimodal,xia2024llmga} that can handle multiple degradations simultaneously using one unified model. While current AiO IR networks are more effective than traditional ones designed for single degradations, training them is more challenging due to the potential conflicts among different optimization objectives \cite{jiang2024survey,chen2024restoreagent,agenticir}. Moreover, these models are typically limited to 3-5 specific tasks seen during training and struggle to generalize to unseen ones. When encountering degradations not included in training, they either require retraining or suffer from a large drop in the quality of recovered images. This hampers their practical application in real-world scenarios, where images are often affected by a wider range of degradations.

The success of large language model (LLM) \cite{achiam2023gpt,touvron2023llama,xi2025rise}-based autonomous AI agents \cite{shen2024hugginggpt,wu2023visual,park2023generative} in handling complex tasks has inspired researchers \cite{chen2024restoreagent,agenticir} to develop intelligent systems to improve the practical applicability of IR methods. In general, agentic systems intelligently perceive degradations in the given LQ image and invoke a series of off-the-shelf pretrained IR networks (referred to as ``tools'') to reverse multiple degradations. They reflect on each step's output and can roll back to previous results to explore more effective tool execution plans. This methodology expands the potential for IR performance improvement while offering greater flexibility than most existing AiO networks. For instance, RestoreAgent \cite{chen2024restoreagent} finetunes a multi-modal LLM (MLLM) \cite{touvron2023llama} to serve as the perception and planning model of an agent, enabling it to solve complex IR problems step-by-step. AgenticIR \cite{agenticir} incorporates statistical experience from pre-collected effective tool execution sequences into LLM \cite{hurst2024gpt}'s text prompts to guide agent in planning, leading to improved quality and consistency of recovered results.

Despite holding potential for autonomous and intelligent IR, existing agentic approaches \cite{chen2024restoreagent,agenticir} still suffer from issues in performance and efficiency, as shown in Fig.~\ref{fig:teaser} (a). \textit{\textbf{First}}, given an LQ input, they search for an effective execution plan in a vast space of tool sequences without considering the characteristics of real-world degradation processes, leading to high resource consumption due to excessive trials and rollbacks with a large number of tool/LLM invocations. For example, AgenticIR can require up to 200 seconds and 20 invocations to restore a $256\times 256$ input image with two NVIDIA 3090 GPUs. \textit{\textbf{Second}}, they rely on a single agent to handle all of perception, planning, invoking tools for degradations, \etc This often results in suboptimal tool execution plans due to the limited capability of one single agent, constraining their effectiveness and practical applicability.

To improve performance while reducing resource consumption, we propose \textbf{MAIR}, a novel \textbf{M}ulti-\textbf{A}gent system for complex \textbf{IR} problems. Moving beyond the single-agent designs in existing methods, MAIR establishes a real-world degradation prior-augmented multi-agent system. The prior is based upon our induction of real-world degradation processes and statistics of effective tool execution plans. As illustrated in Fig.~\ref{fig:teaser} (b), we categorize degradations into three types, and assume that they occur sequentially in most real-world cases: \textbf{(1)} degradations in the scene (\eg, low light and rain) \cite{tarel2009fast,wang2020dcsfn,nan2014bayesian}, \textbf{(2)} degradations introduced by the imaging process (\eg, noise and blur) \cite{dong2015image,mastin1985adaptive,zhang2022deep}, and \textbf{(3)} degradations caused by post-processings (\eg, JPEG compression) \cite{wallace1991jpeg,marcellin2000overview}. Based on this, we propose a three-stage framework that reverses these degradations in the order opposite to their occurrences, effectively reducing the search space and accelerating execution compared to previous approaches that lack the prior. To overcome the performance limitations of single agent, our design incorporates multiple collaborative agents at two levels for more effective IR problem-solving: a ``scheduler'' agent at the first level controls the overall IR process, while multiple ``expert'' agents at the second level leverage tools to address specific single degradations. All tools are registered in our MAIR system using textual descriptions, allowing users to easily add or modify them, and flexibly control the execution process and recovered result using instructions. This agentic system design enables MAIR to perform IR in an autonomous manner more effectively and efficiently, particularly when handling real-world LQ inputs. In summary, our contributions are:

\vspace{3pt}
\noindent \ding{113} (1) We propose a novel agentic IR system that consists of a three-stage framework and a two-level multi-agent design.

\vspace{3pt}
\noindent \ding{113} (2) We develop a three-stage restoration framework based on our proposed real-world image degradation prior.

\vspace{3pt}
\noindent \ding{113} (3) We develop a two-level multi-agent design that consists of a ``scheduler'' agent for overall perception and planning, and multiple ``expert'' agents for degradation removal.

\vspace{3pt}
\noindent \ding{113} (4) Experiments manifest that MAIR achieves competitive image quality and superior efficiency than the previous agentic method \cite{agenticir}. In addition, it enjoys flexibility in following instruction and extensibility in adding new tools.

\section{Related Work}
\label{sec:related_work}
\textbf{Image Restoration (IR)} aims to reconstruct original HQ images from their LQ observations. In the past, a large number of methods have focused on solving single-degradation problems, such as denoising \cite{wu2020unpaired,zhang2022idr}, deraining \cite{gu2024networks,guo2023sky}, dehazing \cite{engin2018cycle,song2023vision,he2010single}, and super-resolution \cite{chen2024adversarial,wu2024seesr,wang2021real,yu2024scaling}. These methods have generally achieved state-of-the-art performance for specific types of degradations. However, real-world images can often suffer from multiple mixed degradations, causing these methods to perform poorly in the scenarios beyond their intended scope. Recent research has explored AiO IR methods \cite{park2023all,li2022all,jiang2024autodir,potlapalli2024promptir,conde2024instructir,kong2024towards,luo2023controlling,ai2024multimodal,xia2024llmga}, aiming to develop a unified framework capable of handling multiple degradations. For example, AirNet \cite{li2022all} employs contrastive learning to help the network distinguish image features between different IR tasks and apply the most appropriate processing. PromptIR \cite{potlapalli2024promptir} and InstructIR \cite{conde2024instructir} introduce additional degradation context to guide the restoration model. MiOIR \cite{kong2024towards} incorporates sequential and prompt learning strategies to enable the network to incrementally learn individual IR tasks in an organized manner. AutoDIR \cite{jiang2024autodir} automatically detects and removes degradations step-by-step. DA-CLIP \cite{luo2023controlling} integrates a pre-trained CLIP model \cite{radford2021learning} within a restoration network to enhance image quality. Despite these advancements, existing AiO approaches still struggle with multi-task learning, making it difficult to balance generalization ability and reconstruction performance.

\textbf{Autonomous Agents} are systems developed to perceive environment, make decisions, and execute actions independently. In recent years, a growing body of research has explored the use of LLMs as core controllers in autonomous agents \cite{yang2023auto,shen2024hugginggpt,wu2023visual,yang2024gpt4tools}. To enhance the ability of AI systems to solve complex problems, researchers have designed various multi-agent frameworks \cite{kim2023llm,hong2023metagpt,wu2023autogen,wang2307unleashing,hao2023chatllm} that enable agents to specialize, coordinate, and collaborate. Some works \cite{park2023generative,zhuge2023mindstorms,yang2024oasis} simulate sociological dynamics utilizing multiple agents, further expanding their capabilities.

A series of notable approaches have emerged in this field. For instance, MetaGPT \cite{hong2023metagpt} introduces human role structures into multi-agent systems, assigning different responsibilities to agents in software development tasks. CAMEL \cite{li2023camel} proposes a role-playing framework, demonstrating the power of structured agent interactions in problem-solving. AutoGen \cite{wu2023autogen} enhances collaborative agentic AI systems by automating agent communications and coordination, while AutoAgents \cite{chen2023autoagents} dynamically generates and organizes specialized agents into teams tailored for various specific tasks.

This evolution of agents has inspired new approaches for IR. RestoreAgent \cite{chen2024restoreagent} finetunes an MLLM \cite{touvron2023llama} as the perception and planning model of agent on synthetic datasets, enabling autonomous evaluation and tool execution. AgenticIR \cite{agenticir} develops a system to mimic a human user, leveraging statistical experience for complex IR tasks. However, the performance and efficiency of existing agentic IR methods are constrained due to the limited capability of single-agent systems and their resource-intensive search for effective tool execution plans. To tackle these problems, we propose a three-stage restoration framework based on our real-world degradation prior to reduce the search space of plans. Additionally, a two-level multi-agent system is introduced, with specialized agents for perception, planning, and reflection, as well as handling individual degradations, improving performance while suppressing total resource consumption.

\section{Method}
\subsection{Problem Definition}
\label{sec:problem_definition}
The task of IR addressed in this work is to recover the original HQ image $\xhq$ from its LQ input image $\xlq$, which undergoes a complex degradation process $\D$, assumed to be a composition of $n$ single degradations $\{\D_1, \D_2, \cdots, \D_n\}$ (such as rain, haze, noise, and blur) applied sequentially. To be formal, this degradation process can be formulated as:
\begin{align}
\begin{split}
\xlq &= \D(\xhq) = (\D_n \circ \cdots \circ \D_2 \circ \D_1)(\xhq)\\
&= \D_n(\cdots \D_2(\D_1 (\xhq))).
\end{split}
\end{align}

The goal is to produce a prediction $\xhqhat$ from $\xlq$ that closely approximates $\xhq$ by applying $n$ tools (\ie, IR models) $\{\R_1, \R_2, \cdots, \R_n\}$, assumed to counteract the effects of $\{\D_1, \D_2, \cdots, \D_n\}$ in the reverse order of their application. Formally, this restoration process can be expressed as:
\begin{align}
\begin{split}
\xhqhat &= \R(\xlq) = (\R_1 \circ \R_2 \circ \cdots \circ \R_n)(\xlq)\\
&= \R_1(\R_2(\cdots \R_n(\xlq))).
\end{split}
\label{eq:restoration}
\end{align}

Under the two assumptions about the degradations and restoration tools outlined above, and building on previous works \cite{chen2024restoreagent, agenticir}, the problem in the context of agentic IR is to determine the optimal selection of tools and their execution order (\ie, a plan for their applications). Given the constraints of limited resources, the challenge is to devise and execute this plan in a way that achieves the highest possible image quality in the recovered output, while minimizing computational cost. In the following, we will elaborate on our induced prior regarding real-world degradations and explain how we use this prior and MLLM-based agents to design an autonomous system that simulates a group of humans to collaboratively implement the process in Eq.~(\ref{eq:restoration}).

\begin{figure*}[!t]
\vspace{-17pt}
\centering
\includegraphics[width=\linewidth]{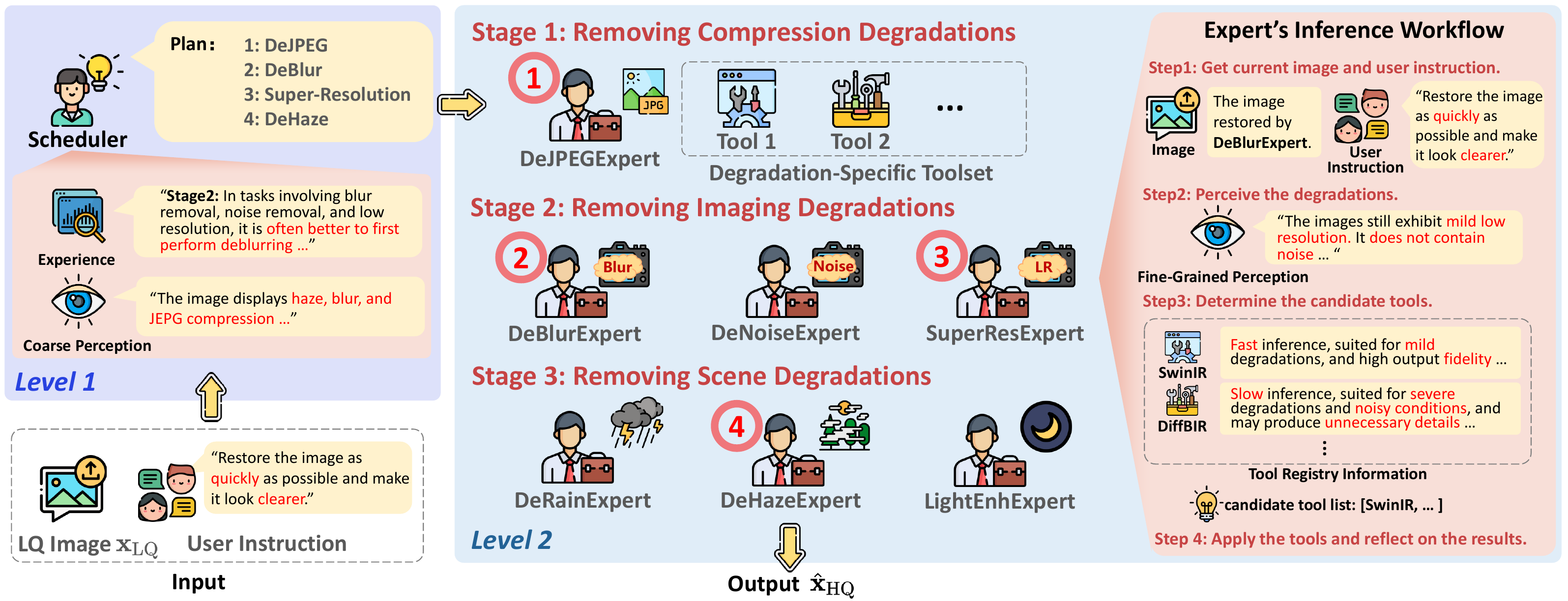}
\vspace{-20pt}
\caption{\textbf{Illustration of the inference workflow of MAIR.} \textbf{(Left)} Given an LQ image and a user instruction, a ``scheduler'' agent first obtains the coarse perception results of degradation types using DepictQA. It then inputs the experience, perception results, and user instruction into GPT-4o to formulate an overall restoration plan, following our three-stage framework. \textbf{(Right)} A group of ``expert'' agents sequentially removes degradations and outputs the reconstructed result, adhering to scheduler's plan. Each expert specializes in a single degradation and uses GPT-4o to intelligently select and apply a list of candidate tools to current image, effectively removing degradation based on the image, instruction, DepictQA's fine-grained perception results of degradation levels, tool registry information, and reflection.}
\label{fig:workflow}
\vspace{-10pt}
\end{figure*}

\subsection{Three-Stage Restoration Framework}
\textbf{Real-World Degradation Prior.} In most real-world cases, image degradation does not occur in all permutations of single degradations but instead generally follows a structured process with an inherent order. This process can be broadly divided into three phases. \textit{\textbf{First}}, the scene itself introduces inherent degradations due to environmental factors such as low light, rain, and haze \cite{he2010single, li2018benchmarking, kong2024towards, li2021low, gu2024networks}. \textit{\textbf{Second}}, during the imaging process, additional degradations arise due to the imperfect propagation of light from the scene to the image sensors. These can include noise, blur, and low resolution, which result from sensor limitations, physical disturbances during capture, and signal processing constraints \cite{zhang2022deep, mastin1985adaptive, dong2015image}. \textbf{\textit{Finally}}, once captured, the image is often post-processed by several information-lossy digital compression techniques like JPEG \cite{wallace1991jpeg} to reduce its storage requirement.

Although the exact ordering of degradations within each phase could vary and some exceptions may exist, the overall degradation often follows a sequence of degradations which can be classified into three types: inherent scene degradation $\D_\text{scene}$, degradation introduced during imaging $\D_\text{imaging}$, and degradation caused by storage compression $\D_\text{compression}$. Later-stage degradations can alter the feature distribution of earlier ones. Formally, this process can be expressed as:  
\begin{equation}
\D = \D_\text{compression}\circ \D_\text{imaging} \circ \D_\text{scene}.
\label{eq:prior}
\end{equation}
Each of $\D_\text{scene}$, $\D_\text{imaging}$, and $\D_\text{compression}$ can be the identity or compositions of single degradations within its respective category. We refer to this structured degradation sequence as \textit{real-world degradation prior}, which extends the previous synthesis pipelines \cite{zhang2021designing,wang2021real} by incorporating scene degradations and broadening the scope from blind super-resolution to general IR tasks. It serves as a simplified guideline for reducing the space of degradation orderings in agentic IR.

\textbf{Three-Stage Framework.} Based on our assumption in Sec.~\ref{sec:problem_definition} that the restoration tools are sufficiently powerful to reverse their corresponding single degradations, the ideal restoration process should follow the inverse of the degradation sequence in Eq.~(\ref{eq:prior}). To be more specific, the restoration should sequentially counteract compression, imaging, and scene degradations. We refer to this approach as \textit{three-stage restoration framework}, which can be formally expressed as:
\begin{equation}
\R =\R_\text{scene}\circ\R_\text{imaging}\circ\R_\text{compression},
\label{eq:three_stage}
\end{equation}
where $\R_\text{scene}$, $\R_\text{imaging}$, and $\R_\text{compression}$ represent either the identity mapping, or the compositions of tool applications to reverse the corresponding three types of degradation, \ie, $\D_\text{scene}$, $\D_\text{imaging}$, and $\D_\text{compression}$, respectively.

\begin{table}[!t]
\centering
\setlength{\tabcolsep}{2pt}
\caption{\textbf{Verification of our degradation prior and three-stage restoration framework} on five real-world validation sets, reporting the probability of the best plans adhering to our framework.}
\vspace{-10pt}
\label{tab:three_stage_val}
\resizebox{\linewidth}{!}{
\begin{tabular}{lccccc}
\toprule
\textbf{Set} & \textbf{T-OLED-Val} & \textbf{RealSR-Val} & \textbf{DRealSR-Val} & \textbf{LHP-/Real-Rain-Val} & \textbf{SIDD-Val} \\
\hline \hline 
Top-1 & 98\% & 88\% & 93\% & 87\% & 91\% \\
Top-3 & 100\% & 100\% & 100\% & 95\% & 100\% \\
\bottomrule
\end{tabular}}
\vspace{-10pt}
\end{table}

To verify the effectiveness of proposed prior and framework, inspired by the approach in \cite{agenticir} for exploring effective plans, we conduct exhaustive restoration attempts using over 13,000 tool execution sequences as plans on our pre-collected five real-world validation image sets. Specifically, we employ the MLLM DepictQA \cite{you2024descriptive} of AgenticIR as the perception model to identify the degradations present in LQ inputs. Based on these perception results, we apply all permutations and combinations of corresponding tools in \cite{agenticir} to reverse the degradations, select the best plans, and check if they align with proposed three-stage framework. To cover a wide range of degradations, the validation datasets include 50 LQ-HQ pairs captured with under-display cameras in T-OLED \cite{zhou2021image} (T-OLED-Val), 85 and 85 low-resolution LQ-HQ pairs from RealSR \cite{cai2019toward} and DRealSR \cite{wei2020component} (RealSR-Val and DRealSR-Val), 75 and 25 rain-degraded LQ-HQ pairs from LHP-Rain \cite{guo2023sky} and RealRain \cite{li2022toward}, merged into a single set (LHP-/Real-Rain-Val), and 100 noisy LQ-HQ pairs from SIDD \cite{abdelhamed2018high} (SIDD-Val). To evaluate plans, we employ an extended version of the scoring function in \cite{chen2024restoreagent}, which aggregates multiple image quality assessment (IQA) metrics, including PSNR, SSIM \cite{wang2004image}, LPIPS \cite{zhang2018unreasonable}, DISTS \cite{ding2020image}, MANIQA \cite{yang2022maniqa}, CLIP-IQA \cite{wang2023exploring}, and MUSIQ \cite{ke2021musiq}. These metrics are standardized and summed as in \cite{chen2024restoreagent} to compute an overall score for each recovered image, reflecting the effectiveness of a plan in restoring each given LQ input, with higher scores indicating better restoration performance.

Tab.~\ref{tab:three_stage_val} presents the statistical results of best plans aligning with our framework across the five sets. We can observe that the top-1 plans follow our framework with a probability of at least 87\%, while the top-3 plans exhibit an even higher probability exceeding 95\%. These findings provide strong empirical support for our prior and three-stage framework, demonstrating their applicability to real-world scenarios.

\begin{figure*}[!t]
\vspace{-15pt}
\setlength{\tabcolsep}{0.5pt}
\centering
\footnotesize
\resizebox{\linewidth}{!}{
\begin{tabular}{ccccccccc}
\includegraphics[width=0.11\linewidth]{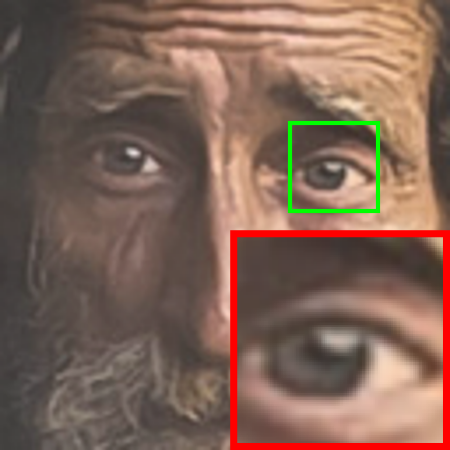}
&\includegraphics[width=0.11\linewidth]{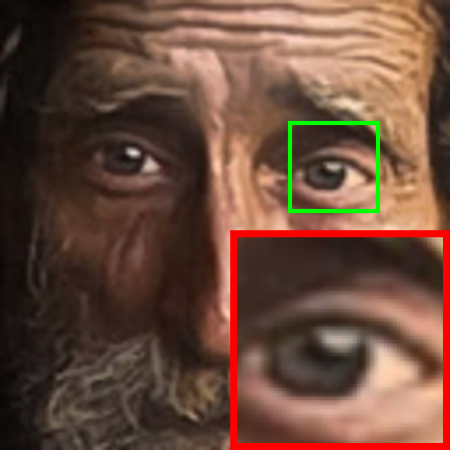}
&\includegraphics[width=0.11\linewidth]{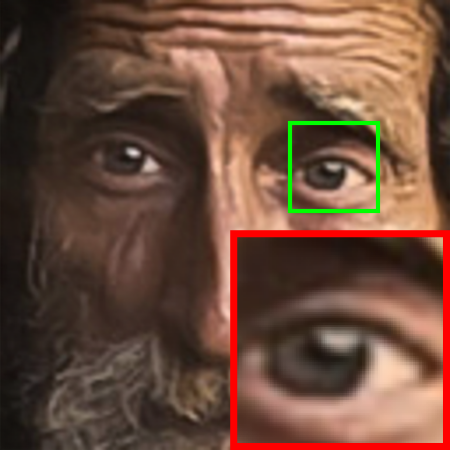}
&\includegraphics[width=0.11\linewidth]{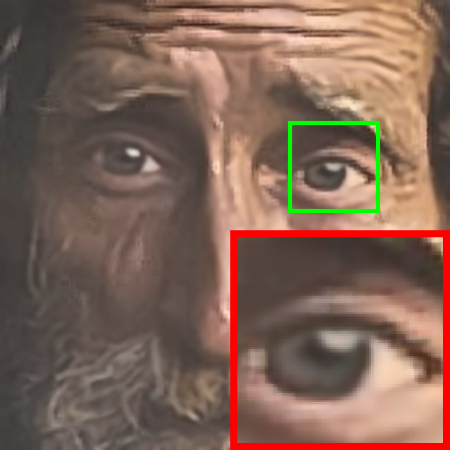}
&\includegraphics[width=0.11\linewidth]{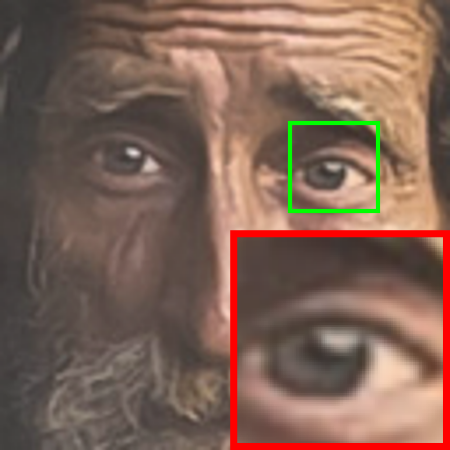}
&\includegraphics[width=0.11\linewidth]{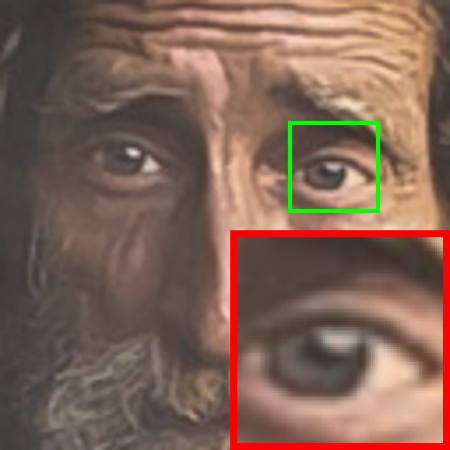}
&\includegraphics[width=0.11\linewidth]{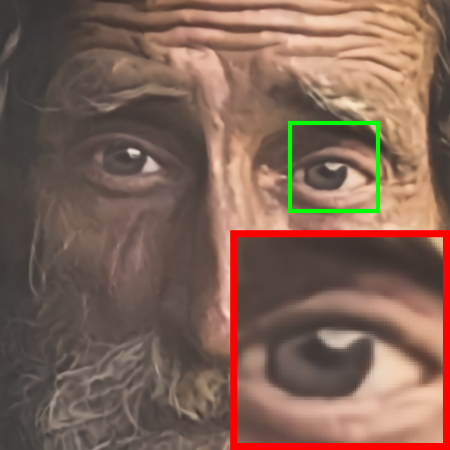}
&\includegraphics[width=0.11\linewidth]{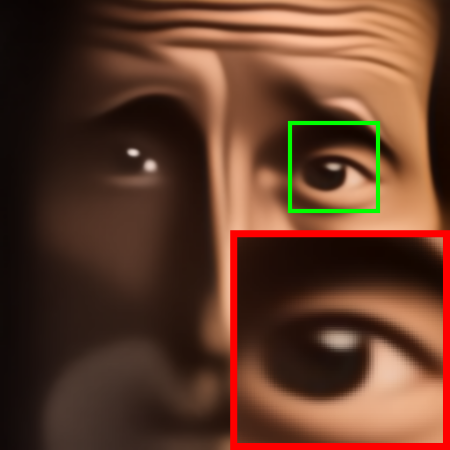}
&\includegraphics[width=0.11\linewidth]{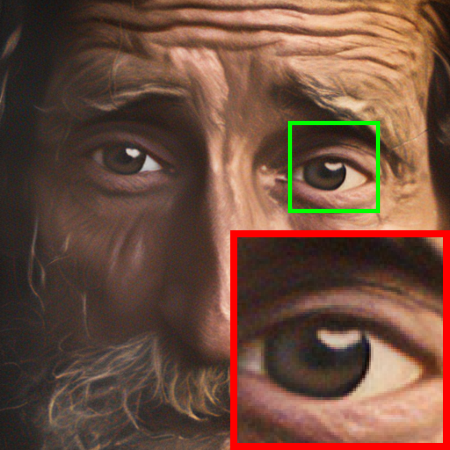}\\
\includegraphics[width=0.11\linewidth]{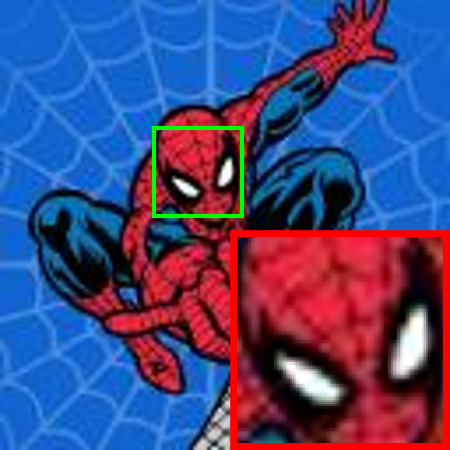}
&\includegraphics[width=0.11\linewidth]{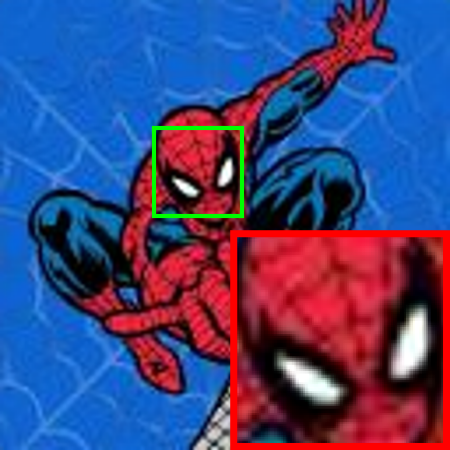}
&\includegraphics[width=0.11\linewidth]{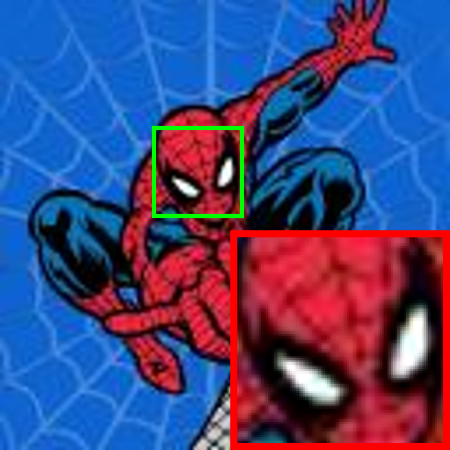}
&\includegraphics[width=0.11\linewidth]{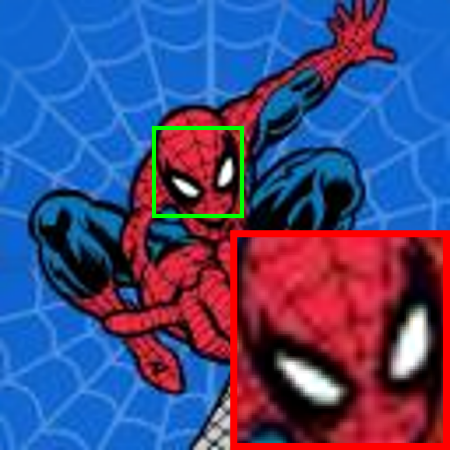}
&\includegraphics[width=0.11\linewidth]{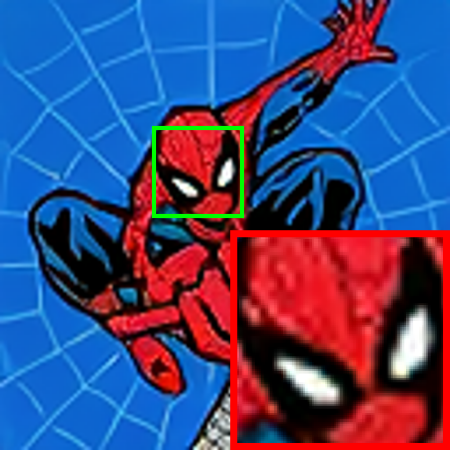}
&\includegraphics[width=0.11\linewidth]{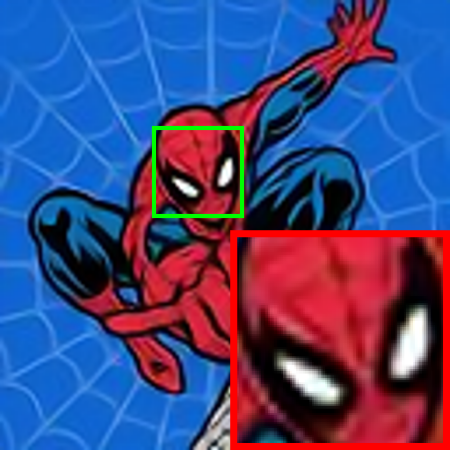}
&\includegraphics[width=0.11\linewidth]{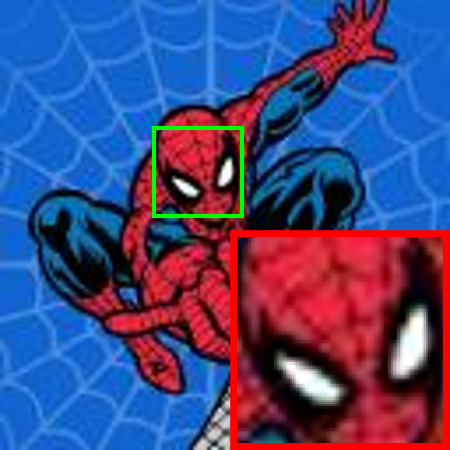}
&\includegraphics[width=0.11\linewidth]{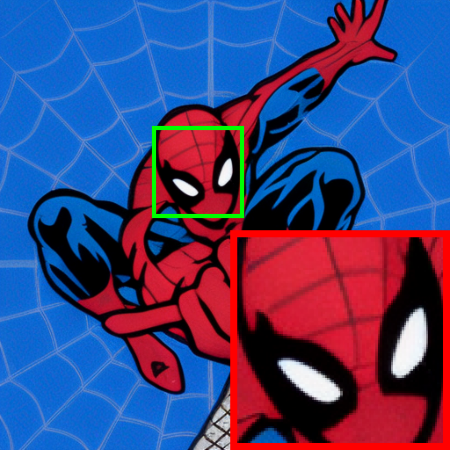}
&\includegraphics[width=0.11\linewidth]{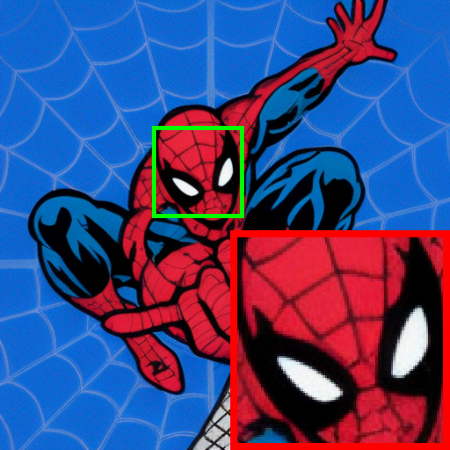}\\
\includegraphics[width=0.11\linewidth]{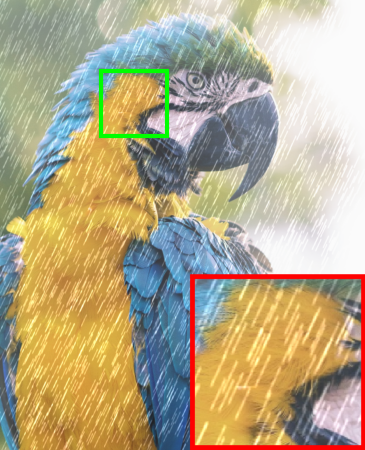}
&\includegraphics[width=0.11\linewidth]{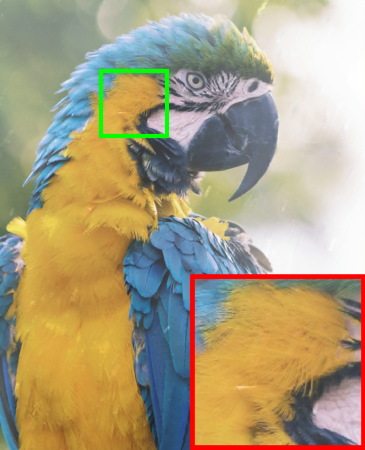}
&\includegraphics[width=0.11\linewidth]{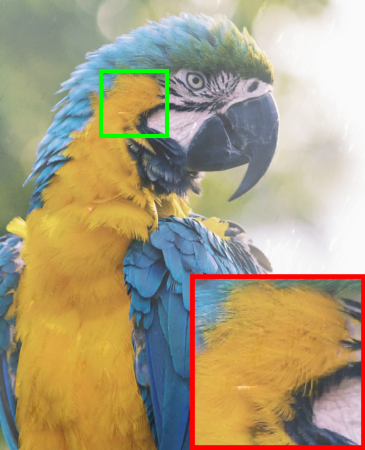}
&\includegraphics[width=0.11\linewidth]{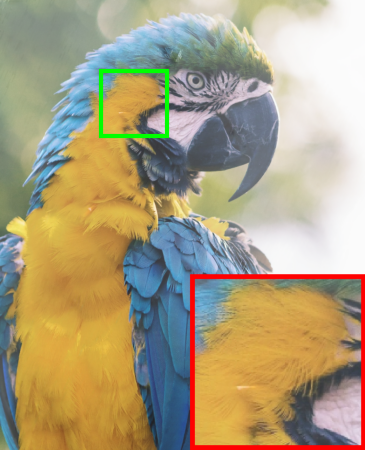}
&\includegraphics[width=0.11\linewidth]{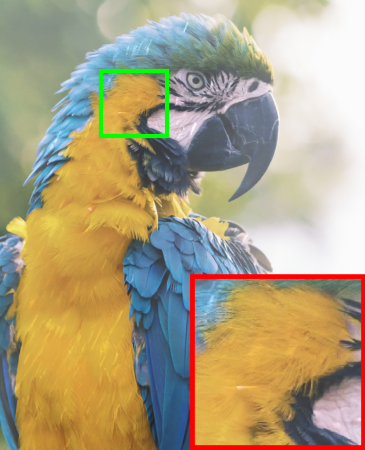}
&\includegraphics[width=0.11\linewidth]{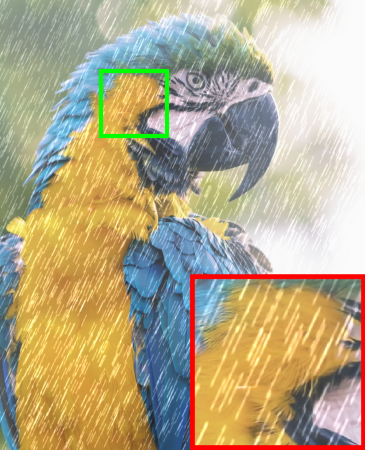}
&\includegraphics[width=0.11\linewidth]{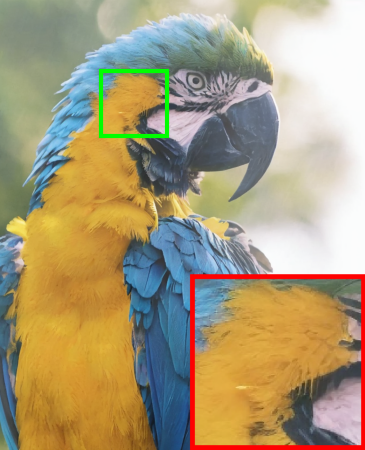}
&\includegraphics[width=0.11\linewidth]{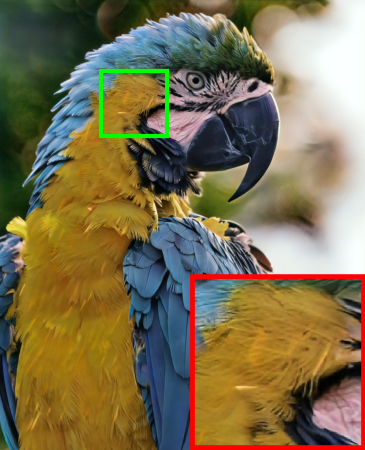}
&\includegraphics[width=0.11\linewidth]{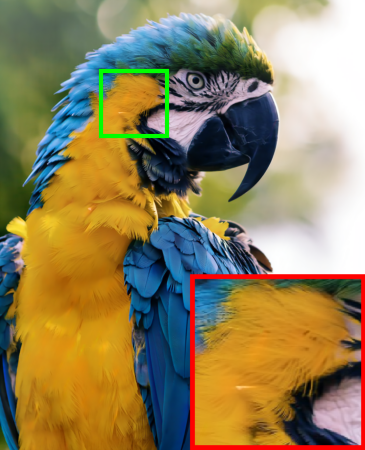}\\
Input & AirNet & PromptIR & MiOIR & DA-CLIP & InstructIR & AutoDIR & AgenticIR & \textbf{MAIR (Ours)} \\
\end{tabular}}
\vspace{-10pt}
\caption{\textbf{Qualitative comparison} on three images from real-world paired (top), unpaired (middle) datasets, and Group A \cite{agenticir} (bottom).}
\label{fig:comparison_qualitative}
\vspace{-10pt}
\end{figure*}

\subsection{Two-Level Multi-Agent System Design}
\label{sec:two_level_multi_agent_design}
\textbf{Overview.} As illustrated in Fig.~\ref{fig:workflow}, our MAIR adopts a two-level multi-agent design. The restoration process begins at the first level, where a ``scheduler'' agent coarsely perceives single degradations present in the LQ input, and formulates an overall restoration plan based on user instruction to counteract them. This plan is then executed at the second level of MAIR system by a group of ``expert'' agents, each specializing in removing a specific type of degradation. During its turn, each expert first conducts a fine-grained perception of degradation level, selects the most appropriate tools from its toolset according to the user instruction, and applies them to the current image while reflecting on their results. Once its restoration is completed, the updated image is passed to the next expert for further processing. This iterative process continues until all relevant expert agents have finished their restorations, ultimately producing the final recovered result.

\begin{table}[!t]
\setlength{\tabcolsep}{2pt}
\caption{\textbf{Quantitative comparison of different methods on three synthesized sets.} Throughout this paper, the best and second-best results are marked in \best{bold red} and \sbest{underlined blue}, respectively.}
\label{tab:comparison_synthesized}
\vspace{-10pt}
\centering
\resizebox{\linewidth}{!}{%
\begin{tabular}{lccccccc}
\toprule[0.9pt]
\textbf{Dataset} & \textbf{Method} & \textbf{PSNR} & \textbf{SSIM} & \textbf{LPIPS$\downarrow$} & \textbf{MANIQA} & \textbf{CLIP-IQA} & \textbf{MUSIQ} \\
\hline\hline
& AirNet     & 19.13 & 0.6019 & 0.4283 & 0.2581 & 0.3930  & 42.46 \\
& PromptIR   & 20.06 & 0.6088 & 0.4127 & 0.2633 & 0.4013 & 42.62 \\
& MiOIR      & 20.84 & 0.6558 & 0.3715 & 0.2451 & 0.3992 & 47.82 \\
Group A & DA-CLIP    & 19.58 & 0.6032 & 0.4266 & 0.2418 & 0.4139 & 42.51 \\
& InstructIR & 18.03 & 0.5751 & 0.4429 & 0.2660 & 0.3528 & 45.77 \\
& AutoDIR    & 19.64 & 0.6286 & 0.3967 & 0.2500   & 0.3767 & 47.01 \\
\cline{2-8}
& AgenticIR  & \best{21.04} & \textbf{\textcolor{red}{0.6818}} & \sbest{0.3148} & \sbest{0.3071} & \sbest{0.4474} & \sbest{56.88} \\
& \textbf{MAIR (Ours)}      & \sbest{21.02} & \sbest{0.6715} & \textbf{\textcolor{red}{0.2963}} & \textbf{\textcolor{red}{0.3330}} & \textbf{\textcolor{red}{0.4751}} & \textbf{\textcolor{red}{59.19}} \\
\midrule
& AirNet     & 19.31 & 0.6567 & 0.367  & 0.2882 & 0.4274 & 47.88 \\
& PromptIR   & 20.47 & 0.6704 & 0.3370  & 0.2893 & 0.4289 & 48.10 \\
& MiOIR      & \sbest{20.56} & 0.6905 & 0.3243 & 0.2638 & 0.4330 & 51.87 \\
Group B & DA-CLIP    & 18.56 & 0.5946 & 0.4405 & 0.2435 & 0.4154 & 43.70  \\
& InstructIR & 18.34 & 0.6235 & 0.4072 & 0.3022 & 0.3790 & 50.94 \\
& AutoDIR    & 19.9  & 0.6643 & 0.3542 & 0.2534 & 0.3986 & 49.64 \\
\cline{2-8}
& AgenticIR  & 20.55 & \textbf{\textcolor{red}{0.7009}} & \sbest{0.3072} & \sbest{0.3204} & \sbest{0.4648} & \sbest{57.57} \\
& \textbf{MAIR (Ours)}       & \textbf{\textcolor{red}{20.92}} & \sbest{0.7004} & \textbf{\textcolor{red}{0.2788}} & \textbf{\textcolor{red}{0.3544}} & \textbf{\textcolor{red}{0.5084}} & \textbf{\textcolor{red}{60.98}} \\
\midrule
& AirNet     & 17.95 & 0.5145 & 0.5782 & 0.1854 & 0.3113 & 30.12 \\
& PromptIR   & 18.51 & 0.5166 & 0.5756 & 0.1906 & 0.3104 & 27.91 \\
& MiOIR      & 15.63 & 0.4896 & 0.5376 & 0.1717 & 0.2891 & 37.95 \\
Group C  & DA-CLIP    & 18.53 & 0.5320  & 0.5335 & 0.1916 & 0.3476 & 33.87 \\
& InstructIR & 17.09 & 0.5135 & 0.5582 & 0.1732 & 0.2537 & 33.69 \\
& AutoDIR    & 18.61 & 0.5443 & 0.5019 & 0.2045 & 0.2939 & 37.86 \\
\cline{2-8}
& AgenticIR  & \sbest{18.82} & \sbest{0.5474} & \sbest{0.4493} & \sbest{0.2698} & \sbest{0.3948} & \sbest{48.68} \\
& \textbf{MAIR (Ours)}       & \textbf{\textcolor{red}{19.42}} & \textbf{\textcolor{red}{0.5544}} & \textbf{\textcolor{red}{0.4142}} & \textbf{\textcolor{red}{0.2798}} & \textbf{\textcolor{red}{0.4239}} & \textbf{\textcolor{red}{51.36}} \\
\bottomrule
\end{tabular}}
\vspace{-15pt}
\end{table}

\textbf{``Scheduler'' for Overall Perception and Planning.} At the first level, a scheduler employs a perception model DepictQA \cite{you2024descriptive} (a finetuned MLLM Vicuna-v1.5-7B \cite{chiang2023vicuna}), as in \cite{agenticir}, to perceive the degradations present in the LQ image. As shown in Fig.~\ref{fig:workflow} (left), it then processes the LQ image, user instruction, and coarse perception results (in text form) using MLLM GPT-4o \cite{hurst2024gpt} to generate a plan that follows our three-stage framework and adheres to the instruction to meet user's specific needs. The plan is an execution sequence of expert agents that reverse single degradations.

One challenge in the planning of scheduler is that while restoration is constrained by our three-stage framework described in Eq.~(\ref{eq:three_stage}), the optimal ordering of single degradation reversions within each stage is unknown without additional priors. For example, noise and blur can coexist in an image, but their optimal reversion order in Stage 2 is uncertain without further guidance. A straightforward approach would be to rely solely on the internal knowledge of GPT-4o. However, this can result in suboptimal plans, as GPT-4o lacks specialized knowledge about intra-stage orderings.

\begin{table}[!t]
\setlength{\tabcolsep}{1pt}
\centering
\caption{\textbf{Quantitative comparison} on real-world paired dataset.}
\vspace{-10pt}
\label{tab:comparison_paired}
\resizebox{\linewidth}{!}{
\begin{tabular}{lcccccccc}
\toprule
\textbf{Method} & \textbf{PSNR} & \textbf{SSIM} & \textbf{LPIPS$\downarrow$} & \textbf{DISTS$\downarrow$} & \textbf{NIQE$\downarrow$} & \textbf{MANIQA} & \textbf{CLIP-IQA} & \textbf{MUSIQ} \\
\hline\hline
AirNet & 22.24 & \sbest{0.7509} & 0.3535 & 0.2256 & 6.59 & 0.2708 & 0.3236 & 42.86 \\
PromptIR & \sbest{24.13} & \best{0.7724} & 0.3337 & \best{0.2159} & 6.49 & 0.2827 & 0.3301 & 42.65 \\
MiOIR & 23.44 & 0.7499 & \best{0.3241} & 0.2231 & 5.61 & 0.2531 & 0.3225 & 44.76 \\
DA-CLIP    & 22.76 & 0.7149 & 0.3694 & 0.2390  & 6.46 & 0.2860  & 0.3126 & 45.24 \\
InstructIR     & \best{26.01} & 0.7910  & 0.3457 & 0.2268 & 6.71 & 0.2882 & 0.3341 & 44.25 \\
AutoDIR    & 20.82 & 0.6770  & 0.3352 & 0.2304 & \sbest{5.49} & \best{0.3329} & 0.3623 & \best{55.74} \\
\hline
AgenticIR    & 19.14 & 0.6574 & 0.3841 & 0.2315 & 5.67 & 0.3152 & \sbest{0.3779} & 52.69 \\
\textbf{MAIR (Ours)}      & 21.67 & 0.7271 & \sbest{0.3244} & \sbest{0.2171} & \best{5.25} & \sbest{0.3199} & \best{0.4030}  & \sbest{55.21} \\
\bottomrule
\end{tabular}}
\vspace{-5pt}
\end{table}

\begin{table}[!t]
\centering
\setlength{\tabcolsep}{2pt}
\caption{\textbf{\fontsize{8.9}{12pt}\selectfont Quantitative comparison} on real-world unpaired dataset.}
\label{tab:comparison_unpaired}
\vspace{-10pt}
\resizebox{0.75\linewidth}{!}{ 
\begin{tabular}{lcccc}
\toprule
\textbf{Method} & \textbf{NIQE$\downarrow$} & \textbf{MANIQA} & \textbf{CLIP-IQA} & \textbf{MUSIQ} \\ \hline \hline
AirNet     & 5.68    & 0.3426  & 0.5175  & 51.16 \\
PromptIR   & 5.89  & 0.3518  & 0.5168  & 51.41 \\
MiOIR      & 6.19  & 0.3677  & 0.5209  & 52.92 \\
DA-CLIP    & 6.48   & \sbest{0.3802}  & 0.5301  & 53.74 \\
InstructIR & 7.02    & 0.3647  & 0.5258  & 56.14   \\
AutoDIR    & 6.32  & 0.3730   & \best{0.5439}  & 53.35 \\ \hline
AgenticIR  & \sbest{5.56}    & 0.3773  & 0.5117  & \sbest{59.13} \\
\textbf{MAIR (Ours)} & \best{5.14} & \best{0.3968} & \sbest{0.5308} & \best{60.08} \\
\bottomrule
\end{tabular}}
\vspace{-5pt}
\end{table}

\begin{table}[t]
\centering
\setlength{\tabcolsep}{5pt}
\caption{\textbf{Efficiency comparison} of average running time and tool invocations per image on real-world paired and unpaired datasets.}
\label{tab:comparison_efficiency}
\vspace{-10pt}
\resizebox{0.6\linewidth}{!}{ 
\begin{tabular}{lccccc}
\toprule
\textbf{Method} & \textbf{Time (s)$\downarrow$} & \textbf{Invocations$\downarrow$} \\
\hline \hline
AgenticIR & 63.04 & 5.15 \\
\textbf{MAIR (Ours)} & \best{35.42} & \best{1.82} \\
\bottomrule
\end{tabular}}
\vspace{-10pt}
\end{table}

\begin{figure*}[!t]
\vspace{-15pt}
\centering
\includegraphics[width=\linewidth]{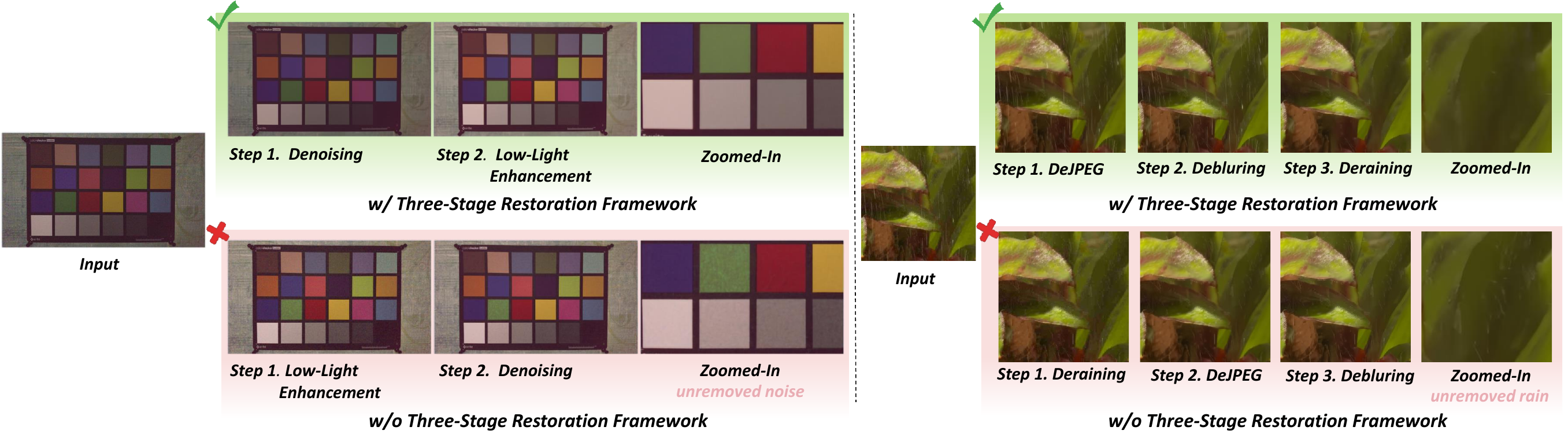}
\vspace{-20pt}
\caption{\textbf{Ablation study of three-stage framework} on two images from real-world paired (left) and LHP-/Real-Rain-Val (right) datasets.}
\label{fig:abla_3stage}
\vspace{-10pt}
\end{figure*}

To fully leverage our attempt records, and the powerful understanding and summarization capabilities of GPT-4o, we adopt an experience-driven technique inspired by \cite{agenticir}. Specifically, we reuse the attempt results from our experiment in Tab.~\ref{tab:three_stage_val}, incorporating both tool execution sequences and corresponding scores. These records are pre-processed offline by GPT-4o in a separate phase to generate text summaries describing typically effective intra-stage restoration orders as \textit{experience} for planning. Since these attempts contain valuable information about the performance of different plans, the extracted ``experience'' serves as a valid guidance input, as shown in Fig.~\ref{fig:workflow} (left), helping the scheduler make more informed decisions when determining the intra-stage restoration order in the inference workflow of our MAIR.

\textbf{``Experts'' for Fine-Grained Degradation Removals.} At the second level, we design experts to recover the image according to the scheduler's plan in a collaborative manner. As shown in Fig.~\ref{fig:workflow} (right), we assign each expert to handle a single degradation (\eg, rain). Concretely, each expert first uses DepictQA to perceive the degradation and its level in the current image (\eg,  high-level additive white Gaussian noise). It then intelligently selects a list of ``candidate'' tools based on the image, user instruction, and the characteristics of its assigned tools. These tools are applied sequentially on the current image, with each result evaluated by DepictQA to check if the degradation has been successfully reduced below the set threshold (\eg, ``low''). If successful, the result is passed to the next agent, and the current agent's task is complete; otherwise, the expert tries other tools from the list. If all candidate tools fail to produce satisfactory results, the expert compares each outcome pairwise and selects the highest-quality image as its restoration output.

\begin{table}[!t]
\centering
\setlength{\tabcolsep}{3pt}
\caption{\textbf{Ablation study of framework} on real-world paired set.}
\label{tab:abla_3stage}
\vspace{-10pt}
\resizebox{\linewidth}{!}{ 
\begin{tabular}{lcccc}
\toprule
\textbf{Method} & \textbf{PSNR} & \textbf{SSIM} & \textbf{LPIPS$\downarrow$} & \textbf{NIQE$\downarrow$}  \\
\hline \hline
w/o Three-Stage Framework & 21.05 & 0.7187 & 0.3256 & 5.30 \\
\textbf{w/ Three-Stage Framework (Ours)}       & \best{21.67} & \best{0.7271} & \best{0.3244} & \best{5.25} \\
\bottomrule
\end{tabular}}
\vspace{-10pt}
\end{table}

\textbf{Tool Registry Mechanism.} Providing the experts with information about tools is crucial for guiding their selection of restoration candidates. While previous agentic methods achieve tool selection and degradation removal by finetuning perception and planning MLLMs (\eg, LLaVA-Llama3-8B) on pairs of LQ images and tool execution sequences \cite{chen2024restoreagent}, or by directly attempting all tools corresponding to a single degradation \cite{agenticir}, they could either require resource-intensive dataset construction and MLLM finetuning, or incur high computational cost due to less intelligence in tool selection. To address this issue, we propose a registry mechanism that equips the experts with specialized knowledge of tools. Specifically, we pre-organize tool information into a set of ``registry'' forms, each corresponding to a specific tool (\eg, SwinIR \cite{liang2021swinir}). These forms record details about all tools' functionality (\eg, denoising), applicable scenarios (\eg, mild levels), efficiency (\eg, fast inference), and other characteristics (\eg, might produce unnecessary details). When an expert agent handles a specific single degradation (\eg, dehazing), the corresponding tool forms are input into GPT-4o as a part of its text input for effective tool selection, eliminating the need for dataset construction and MLLM finetuning, while achieving the flexibility for users to add new tools by simply specifying their corresponding degradation types and filling out the tool registry forms.

Compared to existing single-agent approaches \cite{chen2024restoreagent, agenticir}, as shown in Fig.~\ref{fig:teaser} and our experiments, this two-level multi-agent design with a tool registry mechanism offers superior effectiveness in complex IR problem-solving. This is due to the specialization and coordination of our ``scheduler'' and ``expert'' agents, which decompose the entire problem into manageable sub-tasks, ensuring that both planning and execution are intelligent while preventing any single agent from being overwhelmed. As a result, each agent addresses a focused problem, enhancing both performance and efficiency.

\begin{figure*}[!t]
\vspace{-20pt}
\centering
\centerline{\includegraphics[width=\linewidth]{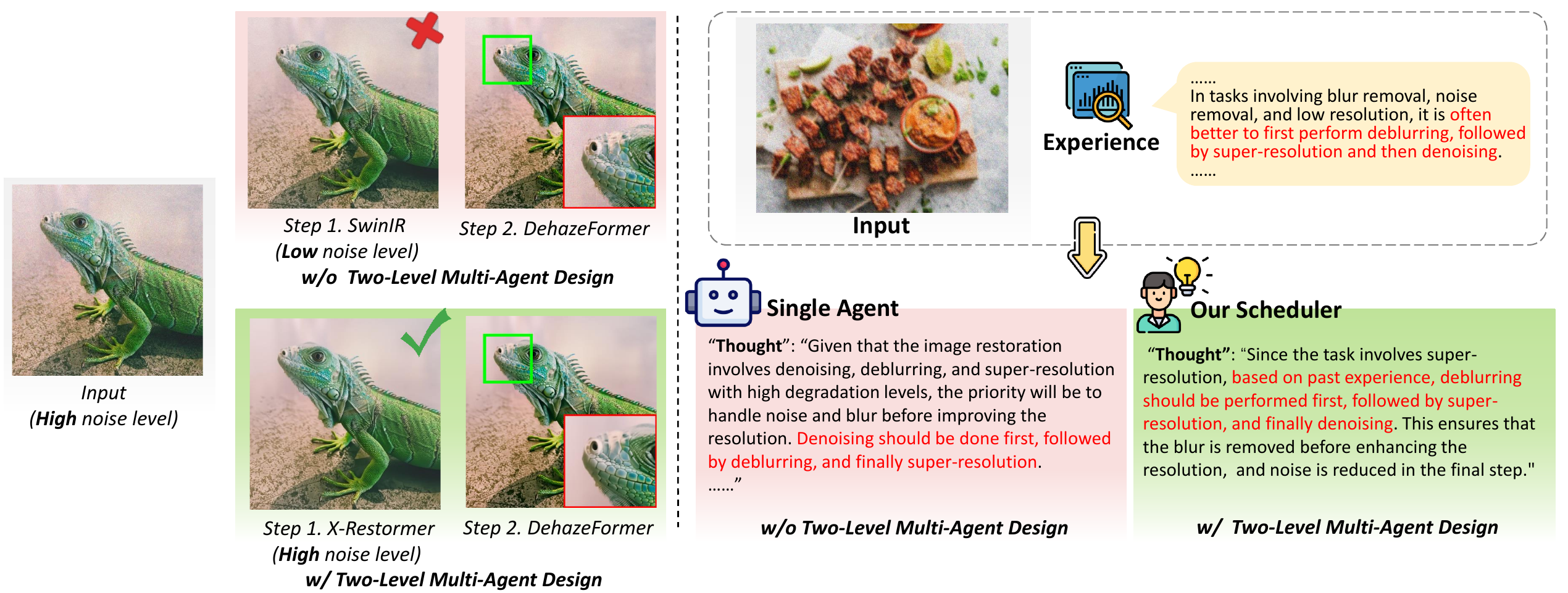}}
\vspace{-5pt}
\caption{\textbf{Ablation study of two-level multi-agent system design} regarding the selection of appropriate tools for effective noise removal on an image from Group B (left), and experience-guided formulation of restoration plans on a synthesized LQ image from MiO100 (right).}
\label{fig:abla_multi_agent_design}
\vspace{-5pt}
\end{figure*}

\begin{table}[!t]
\centering
\setlength{\tabcolsep}{5pt}
\caption{\textbf{Ablation study of the intelligent candidate tool selection of our ``expert'' agents} on the synthetic dataset Group B.}
\label{tab:abla_experts}
\vspace{-10pt}
\resizebox{\linewidth}{!}{ 
\begin{tabular}{lccccc}
\toprule
\textbf{Method} & \textbf{PSNR} & \textbf{LPIPS$\downarrow$} & \textbf{MANIQA} & \textbf{CLIP-IQA} & \textbf{MUSIQ}  \\
\hline\hline
w/o Experts & 20.14 & 0.3427 & 0.2583 & 0.3981 & 50.93 \\
\textbf{w/ Expert (Ours)} & \best{20.92} & \best{0.2788} & \best{0.3544} & \best{0.5084} & \best{60.98} \\
\bottomrule
\end{tabular}}
\vspace{-10pt}
\end{table}

\section{Experiment}
\subsection{Experimental Setting}
\textbf{Implementation Details.} We conduct experiments on both synthesized and real-world test image sets. For synthesized datasets, following \cite{agenticir}, we employ the pre-trained models of SwinIR \cite{liang2021swinir}, FBCNN \cite{jiang2021towards}, DiffBIR \cite{lin2024diffbir}, Restormer \cite{zamir2022restormer}, X-Restormer \cite{chen2024comparative}, DRBNet \cite{ruan2022learning}, DehazeFormer \cite{song2023vision}, RIDCP \cite{wu2023ridcp}, MPRNet \cite{zamir2021multi}, MAIXM \cite{tu2022maxim}, and HAT \cite{chen2023activating} along with traditional operations including gamma correction, constant shift, and histogram equalization as tools for degradation removal. These tools are used to address JPEG artifact removal (Stage 1), denoising, deblurring, and super-resolution (Stage 2), as well as low-light enhancement, deraining, and dehazing (Stage 3). For real-world images, we incorporate RetinexFormer \cite{cai2023retinexformer}, DWGAN \cite{fu2021dwgandiscretewavelettransform}, and CoTF \cite{li2024real} into the toolsets of AgenticIR and MAIR for more effective restoration. The DepictQA model of AgenticIR, and GPT-4o, are used for perception, planning, and reflection. All experiments are conducted on two NVIDIA RTX 3090 GPUs. More details on our MAIR’s workflow, tool models, MLLM inputs, experience summarization, and tool registry forms are provided in the \supp.

\begin{table}[!t]
\vspace{-5pt}
\centering
\setlength{\tabcolsep}{3pt}
\caption{\textbf{\fontsize{8.6}{12pt}\selectfont Ablation study of multi-agent system design} on Group B.}
\label{tab:abla_multi_agent_design}
\vspace{-10pt}
\resizebox{\linewidth}{!}{ 
\begin{tabular}{lccccc}
\toprule
\textbf{Method} & \textbf{PSNR} & \textbf{SSIM} & \textbf{MANIQA} & \textbf{MUSIQ} & \textbf{Tokens $\downarrow$}  \\
\hline \hline
Single-Agent & 20.25 & 0.6612 & 0.3170 & 56.56 & 2410 \\
\textbf{Multi-Agent (Ours)} & \best{20.92}& \best{0.7004} & \best{0.3544} & \best{60.98} & \best{883} \\
\bottomrule
\end{tabular}}
\vspace{-10pt}
\end{table}

\textbf{Test Datasets.} We evaluate MAIR and compare it with other methods using three synthesized test sets: Groups A, B, and C from \cite{agenticir} and two real-world test sets collected by us. The three synthesized test sets contain 1,440 LQ images processed with 16 combinations of mixed 2 or 3 types of degradations applied to images from MiO100 \cite{kong2024towards}. The two real-world test sets consist of real-world samples suffering from multiple unknown degradations. The first set includes 100 paired LQ-HQ image pairs: 10, 10, 15, 15, 10, 20, and 20 pairs from I-Haze \cite{ancuti2018haze}, NH-Haze \cite{ancuti2020nh}, DRealSR \cite{wei2020component}, RealSR \cite{cai2019toward}, T-OLED \cite{zhou2021image}, SIDD \cite{abdelhamed2018high}, and LHP-Rain \cite{guo2023sky}, which do not overlap with the validation sets in Tab.~\ref{tab:three_stage_val}. The second testset contains 100 unpaired real-world LQ images, including 20 hazy images from the internet, 40 images from ImageNet \cite{deng2009imagenet}, and 40 images from RealSR200 \cite{cai2019toward}.

\textbf{Compared Methods} include six AiO IR models: AirNet \cite{park2023all}, PromptIR \cite{potlapalli2024promptir}, MiOIR \cite{kong2024towards}, DA-CLIP \cite{luo2023controlling}, InstructIR \cite{conde2024instructir}, and AutoDIR \cite{jiang2024autodir}, along with AgenticIR \cite{agenticir}.

\textbf{Evaluation Metrics} for assessing the quality of recovered image results include four full-reference IQA metrics: PSNR, SSIM \cite{wang2004image}, LPIPS \cite{zhang2018unreasonable}, and DISTS \cite{ding2020image}, as well as the four no-reference IQA metrics: NIQE \cite{zhang2015feature}, MANIQA \cite{yang2022maniqa}, CLIP-IQA \cite{wang2023exploring}, and MUSIQ \cite{ke2021musiq}. Additionally, FID \cite{heusel2017gans} is also employed to measure the distance between the distributions of ground truth and restored images.

\begin{table}[!t]
\vspace{-5pt}
\centering
\setlength{\tabcolsep}{5pt}
\caption{\textbf{Ablation study of ``experience''} on real-world paired set.}
\label{tab:abla_experience}
\vspace{-10pt}
\resizebox{0.9\linewidth}{!}{
\begin{tabular}{lcccc}
\toprule
\textbf{Method} & \textbf{PSNR} & \textbf{SSIM} & \textbf{LPIPS$\downarrow$} & \textbf{FID$\downarrow$}  \\
\hline\hline
w/o Experience & 20.99 & 0.7041 & 0.3408 & 120.66 \\
\textbf{w/ Experience (Ours)} & \best{21.67} & \best{0.7271} & \best{0.3244} & \best{115.61} \\
\bottomrule
\end{tabular}}
\vspace{-10pt}
\end{table}

\subsection{Comparison with State-of-the-Arts}
\label{sec:comparison_with_sotas}
\textbf{Image Quality Comparison.} As reported in Tabs.~\ref{tab:comparison_synthesized}, \ref{tab:comparison_paired}, and \ref{tab:comparison_unpaired}, MAIR achieves competitive image quality across all five test sets. Specifically, it ranks first or second in Groups A, B, and C, outperforming AgenticIR in all perceptual metrics, including LPIPS, MANIQA, CLIP-IQA, and MUSIQ, while surpassing all the six AiO models. On real-world sets, MAIR delivers competitive results across all metrics, significantly outperforming its baseline agentic approach \cite{agenticir}.

Fig.~\ref{fig:comparison_qualitative} visually demonstrates the superiority of MAIR in restoring three test images. Concretely, it employs a set of tools to effectively remove complex degradations, including haze, noise, low resolution, and compression degradation in the top and middle real-world images, as well as rain and haze in the bottom synthesized image. In contrast, other approaches struggle to reconstruct vivid details around the old man's eye and the anime character's face, or fail to remove JPEG artifacts and haze, leading to suboptimal IR outputs. These results comprehensively validate the effectiveness of proposed MAIR in handling diverse degradation scenarios.

\textbf{Efficiency Comparison.} Tab.~\ref{tab:comparison_efficiency} exhibits that MAIR reduces running time and tool invocations by 44\% and 65\%, respectively, compared to \cite{agenticir}. This efficiency gain is due to our three-stage framework and multi-agent design, which constrain the search space of plans and enable training-free intelligent tool selection, minimizing unnecessary trials and rollbacks while maintaining competitive performance.

\begin{figure}[!t]
\vspace{-20pt}
\centering
\includegraphics[width=\linewidth]{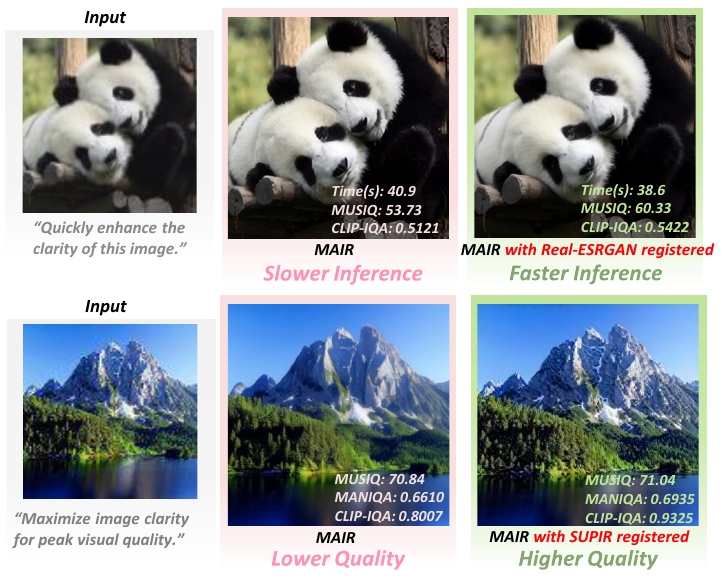}
\vspace{-20pt}
\caption{\textbf{Evaluation of MAIR's extensibility and flexibility in adding new tools} on two images from real-world unpaired dataset.}
\label{fig:extensibility}
\end{figure}

\subsection{Ablation Study}
\textbf{Effect of three-stage framework.} Tab.~\ref{tab:abla_3stage} exhibits that removing the framework from scheduler results in consistent performance drops of 0.62dB in PSNR, 0.0084 in SSIM, 0.0012 in LPIPS, and 0.05 in NIQE. Fig.~\ref{fig:abla_3stage} further demonstrates that without our framework, the scheduler often formulates suboptimal execution orders, which are crucial for removing multiple degradations \cite{chen2024restoreagent,agenticir}. These findings indicate that our proposed prior and framework are effective.

\textbf{Effect of the intelligent tool selection of experts.} Tab.~\ref{tab:abla_experts} exhibits that eliminating intelligent selection (\ie, experts' Step 3 in Fig.~\ref{fig:workflow}) and instead using random tool selections for each degradation leads to performance drops of 0.78 dB in PSNR, 0.0639 in LPIPS, 0.0961 in MANIQA, 0.1103 in CLIP-IQA, and 10.05 in MUSIQ, validating the necessity of expert agents selecting appropriate tools using GPT-4o.

\textbf{Effect of multi-agent system design.} Tab.~\ref{tab:abla_multi_agent_design} shows that replacing multi-agent design with one single agent that handles all model information and experience—aggregated into the input of GPT-4o to address all the sub-tasks—results in image quality drops of 0.67dB in PSNR, 0.0392 in SSIM, 0.0374 in MANIQA, and 4.42 in MUSIQ. Additionally, it increases GPT-4o’s token consumption by 2.73$\times$ due to the aggregation of text descriptions of tool registry forms, experience, \etc, leading to higher costs. Fig.~\ref{fig:abla_multi_agent_design} further manifests that a single-agent approach often selects inappropriate tools and fails to follow the summarized experience, resulting in suboptimal planning and execution, as evidenced by unremoved noise (left) and agents' thoughts (right). These results verify the effectiveness of our multi-agent design in collaboratively solving complex IR tasks while maintaining training-free, intelligent tool selection and flexibility.

\begin{figure}[!t]
\vspace{-20pt}
\centering
\includegraphics[width=\linewidth]{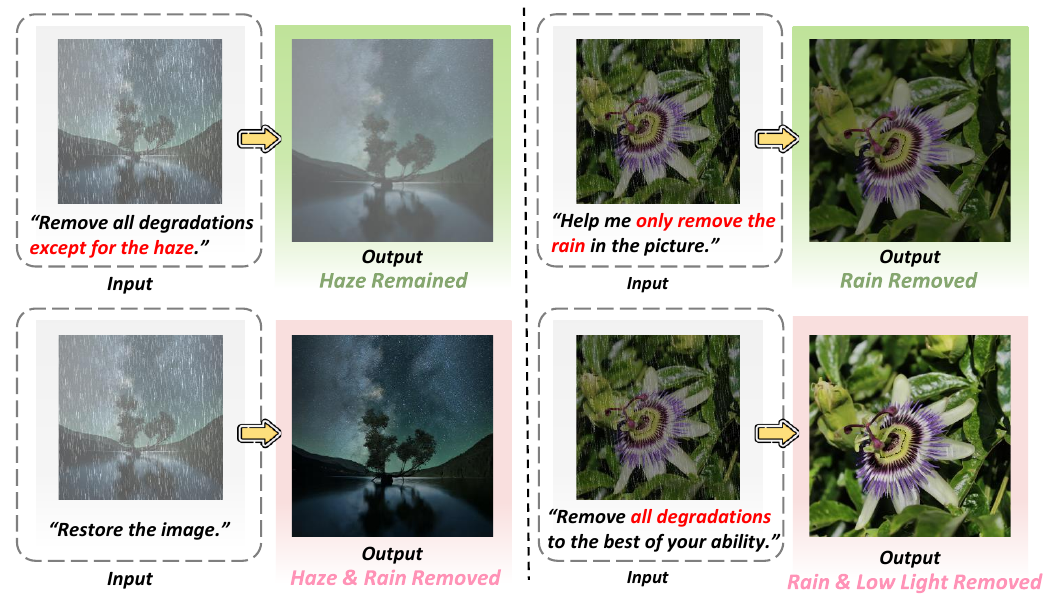}
\vspace{-20pt}
\caption{\textbf{Evaluation of MAIR's instruction-following capability} on two images from the synthetic Groups A (left) and B (right).}
\label{fig:instruction_following}
\vspace{-10pt}
\end{figure}

\textbf{Effect of experience.} Tab.~\ref{tab:abla_experience} exhibits that the elimination of our summarized experience results in performance drops of 0.68dB in PSNR, 0.023 in SSIM, 0.0164 in LPIPS, and 5.05 in FID, highlighting its importance for guiding MAIR in real-world scenarios, as also previously validated in \cite{agenticir}.

\subsection{Evaluation of Extensibility and Flexibility}
Fig.~\ref{fig:extensibility} shows that MAIR allows users to easily extend its capabilities by adding new tools through our registry mechanism. Specifically, we register two models, Real-ESRGAN \cite{wang2021real} (fast inference with moderate quality) and SUPIR \cite{yu2024scaling} (higher quality with lower speed), into super-resolution expert. When users request higher speed or quality, MAIR intelligently calls these models to either reduce running time by 2.3s or enhance details in Stage 2. Fig.~\ref{fig:instruction_following} demonstrates the instruction-following capabilities of MAIR: when users request retaining certain degradations, such as haze or rain, MAIR is able to understand these preferences and perform personalized restoration accordingly. These results comprehensively verify the extensibility, flexibility, and controllability of MAIR—features that are lacking in RestoreAgent \cite{chen2024restoreagent} and AgenticIR \cite{agenticir}, which require MLLM fine-tuning when adding tools or exhibit less intelligence in selection.

\section{Conclusion}  
This paper introduces \textbf{MAIR}, a novel \textbf{M}ulti-\textbf{A}gent system that emulates a team of human specialists to tackle complex \textbf{IR} problems. We model real-world image degradations as a composition of three categories of single degradations and reverse them in the opposite order of their occurrence. Built upon this three-stage framework, we develop a multi-agent system consisting of a ``scheduler'' for planning and multiple ``experts'' specialized in counteracting individual degradations using pre-trained IR models (referred to as ``tools''). A registry mechanism is further introduced to enable easy integration of tools. Experiments on both synthetic and real-world datasets exhibit that proposed MAIR achieves competitive image quality and higher efficiency than the previous agentic IR method \cite{agenticir}, while offering greater flexibility and controllability compared to non-agentic approaches.

Despite being effective, our proposed three-stage framework can not fully cover all real-world degradations, which are often highly complicated and unknown. In addition, although MAIR shows greater flexibility than most AiO models, its inference remains slow, requiring tens of seconds for an image. We leave these limitations for future research.

\small
\bibliographystyle{ieeenat_fullname}
\bibliography{ref}

\end{document}